\documentclass[runningheads]{llncs}
\usepackage{graphicx}
\usepackage{todonotes}
\usepackage{multirow}
\usepackage{booktabs}

\usepackage{array}
\usepackage{tabu}
\newcolumntype{M}[1]{>{\centering\arraybackslash}m{#1}}
\newcolumntype{L}[1]{>{\arraybackslash}m{#1}}

\begin{document}
\title{Overview of BioASQ 2026: The fourteenth BioASQ Challenge on Large-Scale Biomedical Semantic Indexing and Question Answering
}
    \titlerunning{Overview of BioASQ 2026}
\author{
Anastasios Nentidis\inst{1}\orcidID{0000-0002-3782-4412} \and
Georgios Katsimpras\inst{1}\orcidID{0000-0003-3697-941X} \and
Anastasia Krithara\inst{1}\orcidID{0000-0003-0491-4507} \and
Martin Krallinger\inst{2}\orcidID{0000-0002-2646-8782} \and
Miguel Rodríguez-Ortega\inst{2}\orcidID{0009-0000-0188-079X} \and
Eduard Rodríguez-López\inst{2} \and 
Natalia Loukachevitch\inst{3}\orcidID{0000-0002-1883-4121} \and
Igor Rozhkov\inst{3} \and 
Elena Tutubalina\inst{4,5}\orcidID{0000-0001-7936-0284}\and
Dimitris Dimitriadis\inst{6}\orcidID{0000-0002-9404-0331}\and
Vasiliki Patsiou\inst{7}\orcidID{0000-0002-3444-4537}\and
Grigorios Tsoumakas\inst{6,8}\orcidID{0000-0002-7879-669X}\and
George Giannakoulas\inst{6}\orcidID{0000-0001-7491-6319}\and
Alexandra Bekiaridou\inst{7}\orcidID{0000-0002-1143-2293}\and
Athanasios Samaras\inst{6}\orcidID{0000-0002-3404-2749}\and
Giorgio Maria Di Nunzio\inst{9}\orcidID{0000-0001-9709-6392}\and
Nicola Ferro\inst{9}\orcidID{0000-0001-9219-6239}\and
Stefano Marchesin\inst{9}\orcidID{0000-0003-0362-5893}\and
Marco Martinelli\inst{9}\orcidID{0009-0001-1596-8642}\and
Gianmaria Silvello\inst{9}\orcidID{0000-0003-4970-4554}\and
Georgios Paliouras\inst{1}\orcidID{0000-0001-9629-2367}
}

\authorrunning{A. Nentidis et al.}
%
\institute{
National Center for Scientific Research ``Demokritos'', Athens, Greece\\
\email{\{tasosnent, gkatsibras, akrithara,  paliourg\}@iit.demokritos.gr} \and
Barcelona Supercomputing Center, Barcelona, Spain\\
\email{\{martin.krallinger,  mirodrig8,  eduard.rodriguez\}@bsc.es}
\and
Lomonosov Moscow State University, Russia \\
\email{louk\_nat@mail.ru}, \email{fulstocky@gmail.com}
\and
Artificial Intelligence Research Institute, Russia
\and
HSE University, Russia\\
\email{\{tutubalinaev\}@gmail.com} \and
Aristotle University of Thessaloniki, Greece\\
\email{\{dndimitri,greg\}@csd.auth.gr, \{g.giannakoulas, ath.samaras.as\}@gmail.com}
\and
Northwell Health, New Hyde Park, New York, USA; \\
\email{\{spatsiou19,ampekiaridou\}@gmail.com}
\and
Archimedes, Athena Research Center, Greece
\and
University of Padua, Italy\\
\email{\{name.surname\}@unipd.it} 
}
\maketitle              
\begin{abstract}
This paper presents an overview of the fourteenth edition of the BioASQ challenge, organized in the context of the Conference and Labs of the Evaluation Forum (CLEF) 2026. 
BioASQ is an international challenge series that supports progress in biomedical language processing tasks ranging from semantic indexing and information extraction to question answering and summarization.  
In 2026, BioASQ included six shared tasks:
a) \textit{Task 14b} on biomedical semantic question answering. 
b) \textit{Task Synergy14} on question answering for developing biomedical topics.
c) \textit{Task MultiClinSum-2} on multilingual clinical summarization. 
d) \textit{Task BioNNE-R} on extracting relations between nested named entities in Russian and English. 
e) \textit{Task ELCardioCC} on clinical coding in cardiology.
f) \textit{Task GutBrainIE} on gut-brain interplay information extraction.
Across these six tasks, 87 distinct teams participated, submitting more than 1000 runs overall. 
As in previous editions, several submissions reached competitive performance, reflecting the continued progress of state-of-the-art methods across biomedical language processing tasks.     

\keywords{Biomedical knowledge \and Semantic Indexing \and Question Answering}
\end{abstract}
\section{Introduction}
BioASQ was launched in 2012 to advance large-scale semantic indexing and question answering (QA) in biomedicine, and it has gradually expanded to include additional biomedical language processing tasks~\cite{Tsatsaronis2015}.
It pursues this goal through annual shared tasks and benchmark datasets that capture the real information needs of biomedical experts. These tasks include updated versions of established tracks that remain timely, as well as novel tracks introduced to address emerging biomedical information needs.     
Through these shared tasks, research teams worldwide working on clinical coding, information extraction, information retrieval, QA, and summarization in biomedicine gain access to public datasets, a common evaluation infrastructure, and opportunities to exchange ideas through the BioASQ challenges and workshops.

This paper presents a compact overview of the shared tasks, the respective datasets, the participating systems, and their performance in the fourteenth BioASQ challenge, held as a CLEF 2026 lab.
The rest of the paper is structured as follows.
Section~\ref{sec:tasks} first gives a general overview of the shared tasks in BioASQ for 2026, together with the respective datasets.
Section~\ref{sec:participants} then offers a condensed description of participation in each shared task.
More detailed descriptions of systems can be found in the corresponding extended task overviews and in the proceedings of the BioASQ lab.
Next, Section~\ref{sec:results} briefly reports the performance of the systems in each task, using state-of-the-art evaluation measures or manual assessment.
Finally, Section~\ref{sec:conclusion} presents some overall conclusions.

\section{Overview of the tasks}
\label{sec:tasks}

The fourteenth edition of the BioASQ challenge consisted of six tasks~\cite{nentidis2026ecir}:
(i) \textit{Task 14b} on biomedical semantic question answering~\cite{BioASQ2026task14bSynergy}. 
(ii) \textit{Task Synergy 14} on question answering developing biomedical topics~\cite{BioASQ2026task14bSynergy}.
(iii) \textit{Task MultiClinSum-2} on multilingual clinical summarization~\cite{multiclinsum2-overview-2026}. 
(iv) \textit{Task BioNNE-R} on extracting relations between nested named entities in Russian and
English~\cite{bionner-overview-2026}. 
(v) \textit{Task ELCardioCC} on clinical coding in cardiology~\cite{BioASQ2026ElCardioCC}.
(vi) \textit{Task GutBrainIE} on gut-brain interplay information extraction~\cite{BioASQ2026taskGutBrainIE}.
In this section, we describe this year's editions for each of these six tasks with a focus on differences from previous editions of the challenge~\cite{BioASQ2025overview,BioASQ2024overview,nentidis2023results,nentidis2022overview}. 

\subsection{Task 14b}

BioASQ \textit{task 14b} is the fourteenth edition of the established BioASQ \textit{task b} on biomedical QA~\cite{BioASQ2026task14bSynergy}. In 2026, the task was organized in three phases: i) Phase A, where systems received biomedical questions in English and had to retrieve relevant PubMed documents and snippets. ii) Phase A+, where systems had to produce both ``exact'' and ``ideal'' answers. Depending on the question type, the exact answer is either \textit{yes} or \textit{no} (yes/no), a single entity name, such as a disease or gene (factoid), or a set of entity names (list). The ideal answer is a paragraph-length response for all question types. iii) Phase B, where BioASQ experts supplied selected relevant material for each question and systems had to generate answers using this additional evidence. 

For testing, 280 new biomedical questions were annotated with gold documents, snippets, and answers, including both exact and ideal answers.
The training set contained 5,729 biomedical questions from earlier \textit{task b} editions, together with answers and supporting documents and snippets, providing a unique resource for developing biomedical QA systems~\cite{krithara2023bioasq}.
Table \ref{tab:b_data} summarizes the training and test datasets for task 14b.
The test data were divided into four independent bi-weekly batches of 80, 80, 60, and 60 questions, respectively, as shown in Table \ref{tab:b_data}. 

\begin{table}[!htb]
        \caption{Statistics on the training and test datasets of task 14b. The numbers for the documents and snippets refer to averages per question.}\label{tab:b_data}
        \centering
        \begin{tabular}{M{0.09\linewidth}M{0.08\linewidth}M{0.08\linewidth}M{0.09\linewidth}M{0.15\linewidth}M{0.15\linewidth}M{0.15\linewidth}M{0.15\linewidth}}\hline
        \textbf{Batch} 	& \textbf{Size} 	&	\textbf{Yes/No}	&\textbf{List}	&\textbf{Factoid}	&\textbf{Summary}& \textbf{Documents} 	& \textbf{Snippets}  	\\\hline
        Train  & 5729 & 1541  & 1130 & 1695    & 1363    & 10.74   & 13.97      \\
        Test 1         & 80   & 17    & 21   & 23      & 19      & 5.25   & 12.41       \\
        Test 2         & 80   & 21    & 13   & 20      & 26      & 3.64   & 7.98       \\
        Test 3         & 60   & 11    & 17   & 17      & 15      & 3.40   & 6.95       \\
        Test 4         & 60   & 16    & 20   & 11      & 13      & 3.78   & 8.35       \\\hline
        \textbf{Total} & 6009 & 1606  & 1201 & 1766    & 1436    & 10.43   & 13.75    \\\hline
        \end{tabular}
\end{table}

\subsection{Task Synergy 14}

BioASQ \textit{task Synergy} was first introduced in 2021 to encourage research on developing biomedical topics, such as COVID-19~\cite{krithara2021bioasq,Synergy_jamia}.
The task is designed as an ongoing dialogue in which experts pose open-ended questions on developing topics, without knowing in advance whether a definitive answer can already be produced. Systems return relevant material, namely documents and snippets, and the experts assess it. They then provide feedback about relevance and indicate whether the available evidence is sufficient by marking questions as \textit{ready to answer}. The process is repeated across four rounds, with each round considering new material from the evolving document resource~\cite{nentidis2021overview}\footnote{As of 2023, this evolving document resource is PubMed~\cite{nentidis2023results}}.
For questions marked as \textit{ready to answer}, systems also provide exact and ideal answers, which the experts assess, providing feedback that can help systems improve their answers in later rounds. 
Additionally, the experts may mark a question as \textit{closed} when they receive a fully satisfactory answer that is not expected to change, or when the question is no longer of interest.

A training dataset of 400 questions on developing topics, incrementally annotated with relevant material and answers, is available from earlier versions of \textit{task Synergy}~\cite{BioASQ2025task13bSynergy,BioASQ2024task12bSynergy,nentidis2023ceur,nentidis2022ceur,nentidis2021ceur}. 
In \textit{task Synergy 14}, 63 new questions on developing health topics were added, namely cortisol biology and Cushing syndrome, cancer treatment response, molecular and genetic mechanisms, infectious diseases and prophylaxis, diagnostic methods, and mental health topics.
Additionally, 3 questions that remained open from earlier editions were added in round two to be enriched with newer evidence and revised answers~\cite{BioASQ2026task14bSynergy}.
Overall, 66 distinct questions were considered in the four Synergy 14 rounds, which contained 63, 66, 64, and 48 questions, respectively. The corresponding distinct yes/no, list, factoid, and summary counts were 15, 17, 14, and 20 in total.
 
\subsection{Task MultiClinSum-2}

Clinical content, including patient medical records, detailed clinical case reports, and electronic health records (EHRs), continues to grow rapidly across healthcare systems worldwide, covering numerous languages beyond English and reflecting the global scope of medical practice. The extensive length of many clinical reports presents a significant barrier for healthcare professionals who must efficiently extract essential clinical information from these documents. Recent advances in generative and encoder-based transformer models have demonstrated considerable potential for automated summarization, offering the capability to synthesize comprehensive clinical narratives into concise summaries that preserve relevant patient details and clinical findings while significantly reducing document length. Such developments highlight the need to rigorously assess and compare the performance of clinical summarization systems across multiple languages.

Building on the first edition of the task, we present MultiClinSum-2, the second edition of our shared task focusing on automatic summarization of clinical case reports, now expanded to 15 languages: English, French, Spanish, Portuguese, Italian, Russian, Catalan, Norwegian, Danish, Romanian, German, Greek, Dutch, Czech, and Swedish. The task leverages a comprehensive corpus of full clinical cases paired with reference summaries derived from biomedical literature. Table ~\ref{tab:multiclinsum2_dataset_stats} reports the corpus statistics for each language sub-track.



\begin{table}[ht]
\caption{Descriptive statistics per language sub-track of the MultiClinSum-2 dataset. For each language, the number of case-summary pairs and mean word and sentence counts are reported for both full clinical cases and their corresponding reference summaries.}
\centering
\begin{tabular}{lccccc}
\toprule
& & \multicolumn{2}{c}{\textbf{Mean Word Count}} & \multicolumn{2}{c}{\textbf{Mean Sent. Count}}\\
\cmidrule(lr){3-4} \cmidrule(lr){5-6}
\textbf{Language} & \textbf{Nr. Pairs} & \textbf{Full case} & \textbf{Summary} & \textbf{Full case} & \textbf{Summary}\\
\midrule
Catalan (ca)    & 27699 & 604.80 & 118.56 & 27.49 & 5.59 \\
Czech (cs)      & 27488 & 469.89 & 86.37  & 30.78 & 5.73 \\
Danish (da)     & 27652 & 485.14 & 93.76  & 28.75 & 5.70 \\
German (de)     & 27424 & 510.20 & 96.89  & 30.47 & 5.82 \\
Greek (el)      & 26749 & 547.05 & 100.84 & 29.53 & 5.62 \\
English (en)    & 27943 & 528.72 & 98.75  & 28.41 & 5.59 \\
Spanish (es)    & 27743 & 591.95 & 116.16 & 27.57 & 5.59 \\
French (fr)     & 27455 & 615.43 & 119.87 & 27.69 & 5.58 \\
Italian (it)    & 27553 & 572.88 & 113.19 & 27.46 & 5.58 \\
Norwegian (nb)  & 27609 & 485.48 & 95.27  & 28.28 & 5.65 \\
Dutch (nl)      & 27587 & 511.66 & 100.19 & 27.11 & 5.58 \\
Portuguese (pt) & 27593 & 551.87 & 108.86 & 27.22 & 5.58 \\
Romanian (ro)   & 27579 & 551.83 & 110.71 & 26.94 & 5.58 \\
Russian (ru)    & 27086 & 481.95 & 88.55  & 29.66 & 5.58 \\
Swedish (sv)    & 27594 & 459.82 & 88.72  & 27.53 & 5.59 \\
\bottomrule
\end{tabular}
\label{tab:multiclinsum2_dataset_stats}
\end{table}

\subsection{Task BioNNE-R}

In the BioNNE-R Shared Task~\cite{bionner-overview-2026}, we address the challenge of extracting relations between nested named entities, i.e. entities that contain other entities within their boundaries, from  biomedical texts. Most relation extraction datasets are annotated with flat named entities, which cannot contain other named entities. Extraction of relations between nested entities provides additional challenges for models: a relation can connect an uppermost entity and its internal entity, or intersect boundaries of an uppermost entity connecting its internal entity and some external entity. These issues can be problematic even for state-of-the art relation extraction approaches \cite{yandutov2023approaches}. One of the key features of BioNNE-R shared task is the focus on relations between nested entities that are (i) derived from the annotations in the NEREL-BIO corpus~\cite{NERELBIO,NEREL-BIO-COLING-2024} and (ii) supplemented by newly annotated data in both English and Russian. The annotation includes 8 entity types and 14 relation types.  The competition was organized into three subtasks that fell under two evaluation tracks: 1. Two \textbf{monolingual tracks} that treated English and Russian data independently; 2. \textbf{Bilingual track} that required a single bilingual model for the combined Russian and English data. Data statistics for both tracks are summarized in tables~\ref{tab:bionner-data-statistics1} and \ref{tab:bionner-data-statistics2}. All BioNNE-R materials can be found on the shared task's GitHub\footnote{\url{https://github.com/nerel-ds/NEREL-BIO/tree/master/BioNNE-R}} and Codabench pages\footnote{\url{https://www.codabench.org/competitions/13132/}}. 

\begin{table}[!htb]
    \setlength{\tabcolsep}{3.5pt}
    \caption{BioNNE-R Shared Task general document statistics. Entities are annotated entity mentions; relations are gold relation annotations.}
    \centering
    \begin{tabular}{lccc}
    \hline
    \textbf{Split} & \textbf{Docs} & \textbf{Entities} & \textbf{Rels} \\
    \hline
    \textbf{[RU]} \\
    \textbf{train} & 716 & 37,879 & 23,773 \\
    \textbf{dev} & 50 & 3,210 & 2,521 \\
    \textbf{test} & 154 & 9,339 & 7,078 \\
    \hline
    \textbf{[EN]} \\
    \textbf{train} & 55 & 3,861 & 3,193 \\
    \textbf{dev} & 50 & 3,478 & 2,967 \\
    \textbf{test} & 152 & 10,316 & 8,350 \\
    \hline
    \textbf{[Totals]} \\
    \textbf{RU} & 920 & 50,428 & 33,372 \\
    \textbf{EN} & 257 & 17,655 & 14,510 \\
    \textbf{BI} & 1,177 & 68,083 & 47,882 \\
    \\
    \end{tabular}
    \label{tab:bionner-data-statistics1}
\caption{BioNNE-R Shared Task relations and their count in all documents.}
    \begin{tabular}{lccccccccc}
    \hline
    \multirow{2}{*}{Relation} & \multicolumn{4}{c}{Russian} & \multicolumn{4}{c}{English} & \multirow{2}{*}{Total} \\ & train & dev & test & total & train & dev & test & total \\
    \hline 
    \texttt{ABBREVIATION} & 862 & 67 & 244 & 1173 & 97 & 74 & 238 & 409 & 1582 \\
    \texttt{AFFECTS} & 2493 & 288 & 774 & 3555 & 362 & 393 & 920 & 1675 & 5230 \\
    \texttt{ALTERNATIVE\_NAME} & 649 & 97 & 201 & 947 & 95 & 98 & 285 & 478 & 1425 \\
    \texttt{APPLIED\_TO} & 284 & 45 & 41 & 370 & 43 & 54 & 70 & 167 & 537 \\
    \texttt{ASSOCIATED\_WITH} & 3295 & 242 & 987 & 4524 & 325 & 233 & 977 & 1535 & 6059 \\
    \texttt{FINDING\_OF} & 1356 & 85 & 453 & 1894 & 121 & 83 & 529 & 733 & 2627 \\
    \texttt{HAS\_CAUSE} & 2710 & 407 & 790 & 3907 & 579 & 487 & 971 & 2037 & 5944 \\
    \texttt{ORIGINS\_FROM} & 566 & 67 & 135 & 768 & 120 & 79 & 214 & 413 & 1181 \\
    \texttt{PART\_OF} & 1483 & 208 & 502 & 2193 & 204 & 248 & 428 & 880 & 3073 \\
    \texttt{PHYSIOLOGY\_OF} & 1127 & 131 & 313 & 1571 & 150 & 178 & 425 & 753 & 2324 \\
    \texttt{SUBCLASS\_OF} & 7517 & 669 & 2187 & 10373 & 743 & 757 & 2475 & 3975 & 14348 \\
    \texttt{TO\_DETECT\_OR\_STUDY} & 771 & 84 & 239 & 1094 & 164 & 116 & 414 & 694 & 1788 \\
    \texttt{TREATED\_USING}  & 282 & 46 & 61 & 389 & 101 & 88 & 182 & 371 & 760 \\
    \texttt{USED\_IN}  & 378 & 85 & 151 & 614 & 89 & 79 & 222 & 390 & 1004 \\
    \hline
    \textbf{TOTAL} & 23773 & 2521 & 7078 & 33372 & 3193 & 2967 & 8350 & 14510 & 47882 \\
    \hline
    \end{tabular}
    \label{tab:bionner-data-statistics2}
\end{table}

\subsection{Task ELCardioCC}

Cardiovascular diseases affect a significant portion of the global population, accounting for 32\% of global deaths according to WHO\footnote{\url{https://www.who.int/health-topics/cardiovascular-diseases}}. Automated clinical coding plays a crucial role in transforming unstructured real-world medical data gathered from patients into structured information, facilitating clinical research, healthcare analytics, and clinical decision support. This need is particularly important for Greek hospitals, where discharge letters are often stored as free-text clinical narratives, making large-scale analysis, interoperability, and efficient reuse of clinical knowledge challenging. The digitization and structuring of discharge letters through automated ICD-10 coding can therefore substantially improve electronic health record management and support downstream healthcare applications. However, existing research predominantly focuses on English clinical text, leaving other languages, such as Greek, underrepresented. To address this gap, the \textit{ELCardioCC} shared task \cite{BioASQ2025ElCardioCC} focuses on the automatic assignment of cardiology-related ICD-10 codes to discharge letters collected from Greek hospitals.

In detail, participants in the \textit{ELCardioCC} task of 2026 were asked to develop artificial intelligence systems capable of processing Greek cardiology discharge letters and predicting the appropriate ICD-10 diagnostic codes at the document level. To accomplish this goal, participants could employ a wide range of natural language processing and machine learning techniques, including Named Entity Recognition (NER), Entity Linking (EL), and multi-label classification methods. The provided corpus was specifically designed to support experimentation with such approaches, enabling systems to identify clinically relevant concepts and map them to standardized medical knowledge in support of ICD-10 prediction. The training dataset includes 2,500 discharge letters, while the test set comprises 500 letters. System performance was evaluated using the F1 score.

\subsection{Task GutBrainIE}
The scientific literature on the gut-brain axis, which investigates the interplay between the \emph{gut microbiota} and neurological and psychiatric disorders, is rapidly expanding \cite{carabotti2015gut,ghaisas2016gut,appleton2018gut,cryan2020gut}. 
PubMed publications on this topic more than doubled between 2020 and 2025, increasing from approximately 600 to more than 1,500 articles per year, making it increasingly challenging for clinicians and researchers to keep track of new findings.

For this reason, we propose the second edition of the \textit{GutBrainIE} task, which aims to foster the development of Information Extraction (IE) systems able to identify, normalize, and structure relevant biomedical knowledge from scientific publications related to the gut-brain axis and its implications in Parkinson's, Alzheimer's, and Multiple Sclerosis \cite{BioASQ2026taskGutBrainIE}.

Compared with the previous edition, the 2026 edition of \textit{GutBrainIE} introduces two main changes: the dataset has been substantially expanded, from around 1,500 to more than 6,000 annotated documents, and concept-level annotations have been added \cite{BioASQ2025taskGutBrainIE,martinelli-2026-eacl}. Accordingly, we kept two subtasks from the previous edition: Named Entity Recognition (NER), which requires the identification and classification of entity mentions according to the GutBrainIE annotation schema, and Mention-level Relation Extraction (M-RE), which requires the identification and labeling of relations between specific entity mentions. 
We further introduced two new subtasks: Named Entity Recognition and Disambiguation (NERD), which extends NER by requiring each detected entity mention to be linked to a concept identifier from reference biomedical vocabularies, and Concept-level Relation Extraction (C-RE), which evaluates relations at the level of linked concepts rather than surface mentions.

The dataset is organized into Training, Development, and Test sets. 
The Training set is divided into three quality tiers: expert-curated (Gold), student-annotated (Silver), and automatically annotated (Bronze). 
The Development and Test sets contain only expert annotations.

\begin{table}[!htb]
\caption{Dataset statistics for \textit{GutBrainIE}.}
\label{tab:gutbrainie_dataset_summary}
\begin{tabular}{|l|r|r|r|r|r|}
\hline
\textbf{Collection} &
  \multicolumn{1}{c|}{\textbf{\# Docs}} &
  \multicolumn{1}{c|}{\textbf{\# Entities}} &
  \multicolumn{1}{c|}{\textbf{Ents/Doc}} &
  \multicolumn{1}{c|}{\textbf{\# Rels}} &
  \multicolumn{1}{c|}{\textbf{Rels/Doc}} \\ \hline
Train Gold      & 639  & 20530 & 32.13 & 8556  & 13.39 \\
Train Silver    & 1310 & 41409 & 31.61 & 21523 & 16.43 \\
Train Bronze    & 2972 & 89987 & 30.28 & 29692 & 9.99  \\ \hline
Development Set & 80   & 2521  & 31.51 & 1261  & 15.76 \\ \hline
Test Set        & 80   & 2850  & 35.62 & 1285  & 16.06 \\ \hline
\end{tabular}%
\end{table}

\section{Overview of participation}
\label{sec:participants}

Overall, 87 distinct teams took part across the six shared tasks of the fourteenth edition of the BioASQ challenge. Participation by task amounted to 45 teams in task 14b, 7 in Synergy14, 10 in MultiClinSum-2, 7 in BioNNE-R, 6 in ELCardioCC, and 16 in GutBrainIE. Most teams focused on a single task, although four teams submitted to both task 14b and Synergy14.  
This section gives a concise overview of the approaches developed by the participating teams for each BioASQ task. More detailed descriptions will be provided in the extended overview of each task~\cite{BioASQ2026task14bSynergy,multiclinsum2-overview-2026,bionner-overview-2026,BioASQ2026ElCardioCC,BioASQ2026taskGutBrainIE}, and system-specific papers will appear in the
proceedings of the fourteenth BioASQ workshop\footnote{\url{https://www.bioasq.org/workshop2026/proceedings}}.

\subsection{Task 14b}

Task 14b attracted 45 teams, which produced 821 submissions from 151 distinct systems across the four batches and the three phases A, A+, and B. 
Participation remained as high as in the previous edition, with a slightly larger number of submitted systems and runs, showing that the task continues to be timely and relevant~\cite{BioASQ2025overview}.   
Specifically, 34, 28, and 36 teams competed in Phases A, A+, and B, respectively, with 100, 81, and 112 distinct systems and 261, 241, and 319 submissions. Twenty-three teams participated in all three phases. 
As in previous years, the open-source system OAQA~\cite{yang2016learning}, which achieved top performance in older editions of BioASQ~\cite{Krithara2016overview}, was used as a baseline for phase B \textit{exact answers}.

The participating systems explored a broad set of retrieval and generation methodologies. 
Beyond traditional retrieval methods like BM25, systems utilized dense and hybrid retrieval models, which were further refined using techniques such as re-ranking, query-expansion, ontology-guided enhancements and multi-query strategies. 
Most approaches relied on multi-stage Retrieval-Augmented Generation (RAG) pipelines that combined these methods with state-of-the-art Large Language Models (LLMs) (including GPT‑4/5 variants, Gemini, Claude, Llama‑3, BioMistral, OpenBioLLM, Qwen). Furthermore, these systems also integrated various advanced techniques such as few-shot prompting, chain‑of‑thought reasoning, ensemble voting, prompt optimization, and agentic workflows. \cite{Ateia_2026,Panou_26,Liuya_26,Hung_26,Zhao_26,Tsai_26,Avdeiko_26,Volz_26,Nguyen_26,Leekha_26,Wang_26,Shahzadi_26,Antunes_26,Bakker_26,Ribeiro_26,Galat_26,Sims_2026,Kouti_26} 

\subsection{Task Synergy 14}

In BioASQ 14, seven teams participated in task Synergy14. They submitted 79 runs from 18 distinct systems. 
Four of these teams also participated in task 14b, while the remaining three focused on Synergy14. 
The participating systems employed a variety of approaches including hybrid retrieval methods, which combined BM25 with cross-encoder models, task-specific LLMs with evidence-grounded prompting, and RAG architectures that utilized various prompting and optimization techniques \cite{Ateia_2026,Gibson_2026,Sims_2026,Panou_26,Rattanapani_26}.
More detailed descriptions for some of the systems will be available in the workshop proceedings.

\subsection{Task MultiClinSum-2}

In general, participation in the task was very satisfactory, with a broader engagement reflecting the expanded multilingual coverage of this second edition. A total of 9 teams submitted at least one run of their predictions, resulting in 75 runs across all sub-tracks, with each team allowed to submit up to 5 runs per sub-track. As expected, the highest participation was observed in the English sub-track, with 7 teams submitting 20 runs, followed by Spanish (5 teams, 14 runs) and Portuguese (5 teams, 9 runs). Catalan and Dutch also attracted considerable attention, with 4 and 3 teams respectively. Norwegian was the only language that received no submissions, although full coverage was ensured through the developed baseline systems. Table \ref{tab:multiclinsum2_participation} presents the participation received per language sub-track.

\begin{table*}[ht!]
\centering
\caption{Participation overview of MultiClinSum-2 task per team and language subtrack. Numbers indicate runs submitted; --- indicates no participation.}
\label{tab:multiclinsum2_participation}
\resizebox{\textwidth}{!}{%
\begin{tabular}{lcccccccccccccccc}
\toprule
\textbf{Team} & \textbf{EN} & \textbf{ES} & \textbf{FR} & \textbf{PT} & \textbf{IT} & \textbf{RU} & \textbf{CA} & \textbf{NO} & \textbf{DA} & \textbf{RO} & \textbf{DE} & \textbf{EL} & \textbf{NL} & \textbf{CS} & \textbf{SV} & \textbf{Total} \\
\midrule
ixa-sum           & 5 & 5 & 2 & 2 & 2 & --- & 5 & --- & --- & 2 & 2 & --- & 2 & --- & --- & \textbf{27} \\
MediScribes                 & 5 & --- & --- & --- & --- & --- & --- & --- & --- & --- & --- & --- & 5 & --- & --- & \textbf{10} \\
NLP4Health      & 3 & 3 & 3 & 2 & --- & --- & --- & --- & --- & --- & --- & --- & --- & --- & --- & \textbf{11} \\
DACHausa            & 1 & 1 & 1 & 1 & 1 & 1 & 1 & --- & 1 & 1 & 1 & 1 & 1 & 1 & 1 & \textbf{14} \\
InfoLab-FEUP         & 3 & --- & --- & 3 & --- & --- & --- & --- & --- & --- & --- & --- & --- & --- & --- & \textbf{6} \\
PMH-Aqeel           & 1 & 1 & --- & 1 & 1 & 1 & 1 & --- & --- & --- & --- & --- & --- & --- & 1 & \textbf{7} \\
UMUTeam             & --- & 4 & --- & --- & --- & --- & --- & --- & --- & --- & --- & --- & --- & --- & --- & \textbf{4} \\
Aleyna              & 2 & --- & --- & --- & --- & --- & --- & --- & --- & --- & --- & --- & --- & --- & --- & \textbf{2} \\
RaRaMi            & --- & --- & --- & --- & --- & --- & --- & --- & --- & 1 & --- & --- & --- & --- & --- & \textbf{1} \\
\midrule
\textbf{Total runs}      & \textbf{20} & \textbf{14} & \textbf{6} & \textbf{9} & \textbf{4} & \textbf{2} & \textbf{7} & \textbf{0} & \textbf{1} & \textbf{3} & \textbf{3} & \textbf{1} & \textbf{7} & \textbf{1} & \textbf{2} & \textbf{75} \\
\bottomrule
\end{tabular}%
}
\end{table*}

\subsection{Task BioNNE-R}

In total, we have received 20 Codabench registrations for the BioNNE-R task, with 7 teams submitting predictions during evaluation. All seven participated in the English subtask, and five each in the Russian and bilingual subtasks. 

Most teams framed nested relation extraction as a pair-classification problem: candidate entity pairs were enumerated, encoded with typed entity markers inserted at span boundaries, and classified by a fine-tuned transformer with an explicit no-relation class. Encoder choices reflected the bilingual setting, with biomedical English models such as BioBERT \cite{gu2021domain} and BioLinkBERT \cite{linkbert} used for the English subtask and multilingual models such as mDeBERTa-v3 \cite{he2021debertav3}, XLM-RoBERTa \cite{DBLP:journals/corr/abs-1911-02116}, and multilingual BERT \cite{Devlin2018} for the Russian and bilingual subtasks. As the inference-time candidate set is dominated by negative pairs, nearly all teams applied postprocessing to control precision: schema-based filtering of entity-type pairs unseen in training, per-class threshold calibration on the development set, and distance- or sentence-boundary-based candidate pruning. Several teams explored alternatives to the standard classifier: HSE NLP Team used an LLM pipeline to synthesize training examples for rare relation types, aakobiakova fine-tuned an 8B-parameter LLM with LoRA \cite{hu2022lora} and log-probability inference, and LODAC-NII complemented a typed-marker classifier with a document-level relational graph model. More detailed system descriptions are available in the extended task overview~\cite{bionner-overview-2026}.

\subsection{Task ELCardioCC}

The ELCardioCC task engaged six teams, with a total of 19 systems submitted in addition to our baseline model. 

The submitted systems reveal a strong trend toward hybrid and multi-stage architectures that combine Named Entity Recognition (NER), entity linking, and document-level ICD-10 classification. Several teams framed the task as a pipeline problem, first identifying clinically relevant spans and then mapping them to standardized diagnostic codes. Some approaches relied on explicit mention detection followed by concept normalization or classification, often leveraging Greek-specific transformer backbones such as BERT-based Greek models and multilingual variants including XLM-RoBERTa. These systems enriched semantic representations through hierarchical medical knowledge, entity linking strategies, translated terminologies, and external clinical resources. A common motivation behind these designs was to reduce the complexity of the original multi-label document classification task by transforming it into finer-grained mention-level prediction problems, enabling more precise handling of clinical concepts and improving generalization across rare ICD-10 categories.

Another clear trend was the extensive use of ensemble learning, curriculum learning, and specialized optimization strategies to address class imbalance and improve robustness. Multiple teams combined heterogeneous models using weighted voting, routing mechanisms, or metaheuristic search procedures rather than relying on a single architecture. For example, some teams explored sophisticated ensemble frameworks integrating precision-oriented and recall-oriented subnetworks, threshold tuning, asymmetric loss functions, and declarative search strategies for model fusion. These methods highlight the growing emphasis on balancing performance across both frequent and long-tail diagnostic labels, a critical challenge in clinical coding tasks. At the same time, even simpler baselines based on Greek BERT models demonstrated the importance of strong language-specific pretrained transformers as foundational components. Overall, the approaches indicate a broader movement toward combining deep contextual language models with structured clinical knowledge, ensemble optimization, and task decomposition techniques to achieve accurate and scalable automated ICD-10 coding systems for cardiovascular discharge summaries.


\subsection{Task GutBrainIE}
The \textit{GutBrainIE} task registered 18 participating teams. 
Among these, 17 teams participated in NER, 14 in NERD, 11 in M-RE, and 10 in C-RE. 

For NER, most teams adopted token classification approaches based on biomedical pretrained transformer models, including PubMedBERT, BioLinkBERT, BiomedBERT, and BioMedELECTRA, often combined with CRF layers \cite{pubmedbert,linkbert,clark2020electra}. 
Several systems used GLiNER-style models fine-tuned on the GutBrainIE training data \cite{zaratiana-etal-2024-gliner}.
A common strategy was to leverage the quality-based stratification of the training set through filtering, re-weighting, and relabeling of noisier data. 
Many teams also employed ensemble methods, combining models trained with different architectures, hyperparameters, or annotation subsets. 

For NERD, most participants extended their NER pipelines with a Named Entity Linking (NEL) component. 
Common strategies included dictionary-based matching, SapBERT-style biomedical embeddings, nearest-neighbor retrieval, and concept-definition encoders \cite{liu-etal-2021-self,gillick-etal-2019-learning,wu-etal-2020-scalable}.

Across the RE subtasks, participants mainly relied on sequence classification approaches using biomedical pretrained language models finetuned to perform classification of entity pairs marked in-line. 
Common strategies included type-based constraints, negative sampling, and threshold tuning. 
Some teams also experimented with LLMs using prompt- and RAG-based approaches.

\section{Results}
\label{sec:results}
\subsection{Task 14b}

This section reports the evaluation measures and preliminary results for task 14b.
The evaluation in \textit{task 14b} combines expert manual assessment of system responses with automatic scoring using established measures \cite{malakasiotis2020evaluation}, as in earlier editions of the task~\cite{BioASQ2025overview,BioASQ2024overview}. Table~\ref{tab:b_eval} summarizes the official measures by response and question type, based on the preliminary results. 
The final results will become available after the BioASQ experts complete manual assessment and enrich the ground truth with any additional relevant items, answer elements, and synonyms. The online results pages for Phase A\footnote{\url{https://participants-area.bioasq.org/results/14b/phaseA/}}, Phase A+\footnote{\url{https://participants-area.bioasq.org/results/14b/phaseAplus/}}, and Phase B\footnote{\url{https://participants-area.bioasq.org/results/14b/phaseB/}} will be updated when final results are available.

Table {\ref{tab:bA_eval}} reports the batch-level document and snippet retrieval performance of systems participating in Phase A of task 14b. Document retrieval was strongest in the first batch and lower in the later batches, with an average top score across batches of 0.142$\pm0.013$. Top snippet F1 remained in a relatively narrow range across the four batches with an average value of about 0.095$\pm0.008$. Both metrics are lower that respective preliminary results reported in the BioASQ 13, which were 0.323$\pm0.08$ and 0.244$\pm0.021$, respectively~\cite{BioASQ2025overview}.
This can be related to the high percentage of questions from new experts (64\%) in BioASQ 14, who have not contributed to the training dataset.

\begin{table*}[!htb]
\caption{The evaluation measures for \textit{task 14b} per response type and question type~\cite{malakasiotis2020evaluation}.}\label{tab:b_eval}
\centering
\begin{tabular}{L{0.26\linewidth}M{0.21\linewidth}L{0.5\linewidth}}\hline
\textbf{Resp. type (Phase) }               & \textbf{Quest. type} & \textbf{Official measure }                                           \\\hline
Documents  (A)                         & All           & Mean Average Precision (MAP)                                \\\hline
Snippets (A)                            & All           & F1 (based on character overlaps)                     \\\hline
  & List          & F1                                                   \\
Exact ans. (A+ \& B)                                       & Yesno         & macro F1 on ``yes'' \& ``no'' classes               \\
                                       & Factoid       & Mean Reciprocal Rank (MRR)                                  \\\hline
Ideal ans. (A+ \& B)                  & All           & Manual scores for precision, recall, repetition, readability
\\\hline 
\\
\end{tabular}
\caption{The average and top scores of participant systems in Phase A, task 14b. 
 }\label{tab:bA_eval}
    \centering
    \begin{tabular}{M{0.15\linewidth}M{0.2\linewidth}M{0.15\linewidth}M{0.2\linewidth}M{0.15\linewidth}}
    \hline
      & \multicolumn{2}{l}{\textbf{Documents}} & \multicolumn{2}{l}{\textbf{Snippets}} \\\hline
\textbf{Batch }& \textbf{Average MAP}     & \textbf{Top MAP}     & \textbf{Average F1}      & \textbf{Top F1}     \\\hline
1     & 0.163           & 0.283       & 0.041           & 0.091      \\
2     & 0.138           & 0.239       & 0.040           & 0.098      \\
3     & 0.130           & 0.211       & 0.033           & 0.084      \\
4     & 0.137           & 0.231       & 0.044           & 0.106     \\
     \hline     
     \\
    \end{tabular}
\label{tab:bA_res_doc}

\caption{Top scores of participant systems in exact answers, task 14b.}\label{tab:b_exact}
    \centering
    \begin{tabular}{M{0.08\linewidth}M{0.13\linewidth}M{0.13\linewidth}M{0.13\linewidth}M{0.13\linewidth}M{0.13\linewidth}M{0.13\linewidth}}
    \hline
      & \multicolumn{3}{c}{\textbf{Phase B}} & \multicolumn{3}{c}{\textbf{Phase A+}} \\\hline
\textbf{Batch} & \textbf{Factoid MRR} & \textbf{Yes/No macro-F1} & \textbf{List F1} & \textbf{Factoid MRR} & \textbf{Yes/No macro-F1} & \textbf{List F1} \\\hline
1     & 0.522 & 1.000 & 0.361 & 0.478 & 1.000 & 0.234 \\
2     & 0.400 & 1.000 & 0.472 & 0.400 & 0.948 & 0.391 \\
3     & 0.588 & 1.000 & 0.518 & 0.529 & 1.000 & 0.326 \\
4     & 0.576 & 1.000 & 0.654 & 0.409 & 0.935 & 0.484 \\
     \hline
    \end{tabular}
\end{table*}

The top exact-answer performance of the participating systems is summarized in Table~\ref{tab:b_exact} for both Phase B and Phase A+. 
In Phase B, the leading systems reached perfect yes/no macro-F1 in all four batches, while factoid MRR ranged from 0.400 to 0.588 and list F1 from 0.361 to 0.654. 
In Phase A+, where systems were not given expert-selected relevant material, the best yes/no scores remained close to Phase B performance, whereas factoid and list scores were lower overall, highlighting that the retrieval of relevant material is a considerable limitation.

\subsection{Task Synergy 14}

In \textit{task Synergy 14}, the same evaluation measures used for \textit{task 14b} are applied, while the information retrieval component considers only newly retrieved material, following the residual collection evaluation approach~\cite{Salton1990}. 
In addition, due to the developing nature of the topics, no answer is available for all of the open questions in each round. Therefore, only the questions indicated as ``answer ready'' were evaluated for \textit{exact} and \textit{ideal answers} per round.

Table~\ref{tab:syn_data} presents the best performance obtained by participating systems per round in \textit{task Synergy 14}.
In Round 1, we started with 63 new questions; no questions were answer-ready, and no answers were received in this round, only relevant material.  
The four-round dialogue between biomedical experts and QA systems in this task supported the progressive collection of enough relevant material to answer 61 of these 63 new questions (97\%) and three additional questions from the previous version of the task that remained open. 
In particular, for 48 questions (73\%), the systems managed to provide at least one ideal answer, which was considered ground-truth quality by the respective expert. 
Overall, the task aided the generation of \textit{exact} and \textit{ideal} answers to questions on developing topics, including the Cushing syndrome, cancer therapy, molecular genetics, infections and prophylaxis, diagnostics, and mental health.

\begin{table}[!htb]
\caption{The number of questions (Qs) and ``Answer Ready'' (AR) questions and the top system performance per round (R) in Task Synergy14. Retrieval of documents (Top MAP) and snippets (Top F1). Generation of exact factoid (Top MRR), list (Top F1), and yes/no (Top macro-F1) answers.
 }\label{tab:syn_data}
    \centering
    \begin{tabular}{M{0.04\linewidth}M{0.08\linewidth}M{0.06\linewidth}M{0.13\linewidth}M{0.16\linewidth}M{0.13\linewidth}M{0.16\linewidth}M{0.18\linewidth}}
    \hline
\textbf{R} & \textbf{Qs} & \textbf{AR} & \textbf{Top MAP} & \textbf{Top F1 Snip.} & \textbf{Top MRR} & \textbf{Top F1 list} & \textbf{Top macro-F1} \\\hline
1 & 63 & 0  & 0.538 & 0.340 & -     & -     & -     \\
2 & 66 & 42 & 0.337 & 0.262 & 0.545 & 1.000 & 1.000 \\
3 & 64 & 55 & 0.249 & 0.390 & 0.394 & 0.818 & 0.771 \\
4 & 48 & 47 & 0.315 & 0.197 & 0.500 & 0.900 & 0.890 \\
     \hline                 
    \end{tabular}
\end{table}

\subsection{Task MultiClinSum-2}

In this second edition, a new detailed evaluation was designed with the objective of providing a more comprehensive and clinically-oriented summary assessment, going beyond the lexical and semantic metrics used in the first edition. For the submissions evaluation, automatically generated summaries were compared against reference summaries using a combined evaluation framework that integrates automatic summarization metrics with an LLM-as-judge approach, capturing both surface-level lexical overlap and semantic similarity alongside deeper clinical quality dimensions. The evaluation metrics considered are described below.

\begin{itemize}
    \item \textbf{ROUGE-1/ROUGE-2} \cite{Lin2004}: Measurement of unigram and bigram overlap between the generated and reference summaries, providing surface-level assessment of lexical similarity.
    \item \textbf{ROUGE-Lsum} \cite{Lin2004}: A sentence-level variant of the ROUGE-L metric that computes the longest common subsequence (LCS) score at the summary level, offering a measure of content coverage and structural similarity with respect to the reference summary.
    \item \textbf{BERTScore} \cite{zhang2020bertscore}: A semantic similarity metric 
    that leverages contextualized embeddings from pre-trained transformer 
    models to capture meaning beyond surface-level lexical overlap.
    \item \textbf{LLM-as-judge}: An evaluation approach powered by EuroLLM-9B-Instruct model \cite{martins2024eurollmmultilinguallanguagemodels}, aiming to assess generated summaries across four clinical dimensions: Faithfulness (factual consistency), Completeness (coverage of key clinical information), Fluency (grammatical correctness) and Consistency (coherence across language versions). 
\end{itemize}

Participating systems presented a broad methodological diversity, with two main strategy identified. On one side, several teams applied fine-tuning approaches with varying degrees of complexity: some used parameter-efficient methods such as QLoRA \cite{dettmers2023qloraefficientfinetuningquantized} for supervised instruction tuning, others explored reinforcement learning through Group Relative Policy Optimization (GRPO) \cite{shao2024deepseekmathpushinglimitsmathematical} in order to optimize generated summaries quality based on lexical and factual consistency metrics, and one team opted for a continual pretraining step on medical-domain models before applying task-specific instruction tuning. On the other hand, a second group of teams approached the task through inference-only systems, focusing on prompt engineering, dynamic example retrieval, or even the addition of external structured knowledge such as medical ontologies or knowledge graphs.

The evaluation results for each language sub-track are summarized in Table ~\ref{tab:multiclinsum2_leaderboard}, showing the top-2 performing teams per language ranked by BERTScore. Team ixa-sum, with their GRPO-based system, consistently ranked first across most sub-tracks on both lexical and semantic metrics, suggesting a clear advantage over other strategies. Other systems such as NLP4Health and MediScribes achieved notable performance on LLM-as-judge clinical dimensions, particularly \textit{completeness} and \textit{faithfullness}. In terms of lower-resource languages, team DACHausa was the main participating team, providing the only submission in Greek, Czech and Danish. Regarding LLM-as-judge evaluation, \textit{consistency} was the weakest clinical dimension, suggesting the limited capability of submitted systems to generate equivalent summaries across different language versions.

\begin{table}[!htb]
\centering
\caption{Top-2 participant teams per MultiClinSum-2 sub-track, ranked by BERTScore. Developed baseline system results are not included in the ranking (R-1 = ROUGE-1, R-2 = ROUGE-2, R-L = ROUGE-Lsum, Faith. = Faithfulness, Com. = Completeness, Fl. = Fluency, Cons. = Consistency).}
\label{tab:multiclinsum2_leaderboard}
\begin{tabular} {L{0.05\linewidth}M{0.2\linewidth}L{0.07\linewidth}L{0.15\linewidth}L{0.06\linewidth}L{0.1\linewidth}L{0.08\linewidth}L{0.08\linewidth}L{0.08\linewidth}L{0.07\linewidth}L{0.07\linewidth}}
\toprule
Lang. & Team & Run & R-1 (R-2) & R-L & BERT Score & Faith. & Com. & Fl. & Cons. \\
\midrule
\midrule
\multirow{2}{*}{EN} & ixa-sum & 2 & \textbf{0.44 (0.22)} & \textbf{0.32} & \textbf{0.88} & 0.73 & 0.83 & 0.70 & \textbf{0.70} \\
 & NLP4Health & 1 & 0.38 (0.16) & 0.27 & 0.87 & \textbf{0.76} & \textbf{0.94} & \textbf{0.70} & \textbf{0.70} \\
\midrule
\multirow{2}{*}{ES} & ixa-sum & 2 & \textbf{0.47 (0.23)} & \textbf{0.31} & \textbf{0.88} & 0.71 & 0.80 & 0.70 & \textbf{0.65} \\
 & NLP4Health & 1 & 0.41 (0.18) & 0.27 & 0.87 & \textbf{0.72} & \textbf{0.84} & \textbf{0.70} & 0.63 \\
\midrule
\multirow{2}{*}{PT} & NLP4Health & 1 & \textbf{0.38 (0.16)} & \textbf{0.26} & \textbf{0.87} & 0.74 & 0.89 & \textbf{0.70} & 0.64 \\
 & ixa-sum & 1 & 0.38 (0.15) & 0.26 & 0.87 & \textbf{0.74} & \textbf{0.91} & 0.70 & \textbf{0.65} \\
\midrule
\multirow{2}{*}{CA} & ixa-sum & 2 & \textbf{0.46 (0.23)} & \textbf{0.30} & \textbf{0.88} & \textbf{0.71} & \textbf{0.87} & \textbf{0.70} & \textbf{0.63} \\
 & DACHausa & 1 & 0.37 (0.16) & 0.25 & 0.84 & 0.66 & 0.69 & 0.69 & 0.34 \\
\midrule
\multirow{2}{*}{NL} & ixa-sum & 2 & \textbf{0.35 (0.11)} & \textbf{0.24} & \textbf{0.86} & 0.73 & 0.95 & \textbf{0.71} & 0.66 \\
 & MediScribes & 3 & 0.34 (0.11) & 0.22 & 0.86 & \textbf{0.75} & \textbf{0.98} & 0.71 & \textbf{0.67} \\
\midrule
\multirow{2}{*}{FR} & ixa-sum & 2 & \textbf{0.41} (0.17) & 0.26 & \textbf{0.87} & 0.74 & \textbf{0.90} & \textbf{0.70} & \textbf{0.67} \\
 & NLP4Health & 1 & 0.41 (\textbf{0.18}) & \textbf{0.26} & 0.87 & \textbf{0.72} & 0.83 & 0.70 & 0.65 \\
\midrule
\multirow{3}{*}{IT} & ixa-sum & 1 & \textbf{0.36 (0.14)} & \textbf{0.24} & \textbf{0.87} & \textbf{0.72} & \textbf{0.89} & \textbf{0.70} & \textbf{0.65} \\
 & PsychAI-Discovery-Lab & 1 & 0.33 (0.12) & 0.24 & 0.86 & 0.70 & 0.79 & 0.70 & 0.56 \\
\midrule
\multirow{2}{*}{RO} & ixa-sum & 1 & \textbf{0.36} (0.15) & \textbf{0.25} & \textbf{0.87} & \textbf{0.74} & \textbf{0.91} & \textbf{0.70} & \textbf{0.57} \\
 & RaRaMi & 1 & 0.34 (\textbf{0.15}) & 0.24 & 0.86 & 0.70 & 0.68 & 0.68 & 0.30 \\
\midrule
\multirow{2}{*}{DE} & ixa-sum & 1 & \textbf{0.30} (0.09) & \textbf{0.20} & \textbf{0.86} & \textbf{0.72} & \textbf{0.83} & \textbf{0.70} & \textbf{0.65} \\
 & DACHausa & 1 & 0.27 (\textbf{0.10}) & 0.19 & 0.82 & 0.63 & 0.46 & 0.69 & 0.34 \\
\midrule
\multirow{2}{*}{RU} & PsychAI-Discovery-Lab & 1 & 0.29 \textbf{(0.07)} & 0.28 & \textbf{0.86} & \textbf{0.70} & \textbf{0.58} & \textbf{0.70} & \textbf{0.39} \\
 & DACHausa & 1 & \textbf{0.32 (0.07)} & \textbf{0.31} & 0.83 & 0.63 & 0.42 & 0.68 & 0.27 \\
\midrule
\multirow{2}{*}{SV} & PsychAI-Discovery-Lab & 1 & \textbf{0.33 (0.12)} & \textbf{0.23} & \textbf{0.86} & \textbf{0.71} & \textbf{0.78} & 0.70\textbf{} & \textbf{0.61} \\
 & DACHausa & 1 & 0.29 \textbf{(0.12)} & 0.21 & 0.83 & 0.64 & 0.58 & 0.69 & 0.36 \\
\midrule
DA & DACHausa & 1 & \textbf{0.29 (0.12)} & \textbf{0.21} & \textbf{0.83} & \textbf{0.63} & \textbf{0.58} & \textbf{0.68} & \textbf{0.35} \\
\midrule
EL & DACHausa & 1 & \textbf{0.31 (0.07)} & \textbf{0.30} & \textbf{0.85} & \textbf{0.65} & \textbf{0.58} & \textbf{0.69} & \textbf{0.28} \\
\midrule
CS & DACHausa & 1 & \textbf{0.30 (0.10)} & \textbf{0.19} & \textbf{0.83} & \textbf{0.63} & \textbf{0.45} & \textbf{0.68} & \textbf{0.29} \\
\bottomrule
\end{tabular}
\end{table}



\subsection{Task BioNNE-R}

As an official baseline, we provide a relation classifier built on the OpenNRE framework \cite{opennre} with a \texttt{bert-base-multilingual-cased} backbone \cite{Devlin2018}. Each candidate entity pair is encoded with special entity markers inserted around the head and tail spans, and classified into one of the 14 relation types or a \texttt{no\_relation} class. For each candidate pair, the minimal sentence segment surrounding the two entities is extracted as model input. Training data is constructed by enumerating gold entity pairs and adding negative (no\_relation) examples at a 3:1 negative-to-positive ratio, with candidate pairs constrained to entity-type combinations permitted by the annotation schema. Separate checkpoints were trained for the English, Russian, and bilingual tracks, the latter on the concatenation of the English and Russian training data. 

Systems were evaluated against the gold relation annotations using precision, recall, and F1-score. Given the highly imbalanced relation-type distribution, macro-averaged F1 across relation types was adopted as the official ranking metric, computed independently for each subtask.  

\begin{table}[t]
\centering
\caption{Official BioNNE-R results: macro-F1 of each team's best
submission per subtask. Best result per subtask in bold; -- indicates
no submission.}
\label{tab:bionner-results}


\setlength{\tabcolsep}{12pt}

\begin{tabular}{lccc}
\hline
Team & English & Russian & Bilingual \\
\hline
ELiRF-UPV         & \textbf{0.5060} & 0.4717          & 0.4540 \\
LODAC-NII     & 0.5016          & \textbf{0.5283} & \textbf{0.5015} \\
aakobiakova   & 0.4951          & 0.4403          & 0.4716 \\
HSE NLP  & 0.4733          & 0.5057          & 0.4527 \\
rabiaozdemir  & 0.4620          & --              & -- \\
savvafq       & 0.4570          & 0.4829          & 0.4849 \\
LSI\_UNED     & 0.4514          & --              & -- \\
\hline
Baseline      & 0.2825          & 0.3070          & 0.3027 \\
\hline
\end{tabular}
\end{table}

Table~\ref{tab:bionner-results} reports macro-F1 for the best submission of each team. All participating systems substantially outperformed the baseline, yet scores cluster in a narrow 0.44--0.53 band across subtasks, indicating considerable headroom for nested relation extraction. ELiRF-UPV achieved the best English result (0.5060), while LODAC-NII led on both the Russian (0.5283) and bilingual (0.5015) subtasks, with only a 0.0044 margin separating the two top English systems. Despite the much larger Russian training set, Russian and English scores were comparable and among the five teams that entered both subtasks, the mean macro-F1 was 0.4858 for Russian against 0.4866 for English, although the single best Russian score (0.5283) did exceed the best English one. The document-level graph approach and the typed-marker pair-classification approaches performed comparably, with no approach type showing a clear advantage.

\subsection{Task ELCardioCC}

The results of the best systems regarding of F1 score of the participants for the ELCardioCC task are presented in table \ref{tab:elcard}.   

\begin{table}[!ht]
\caption{Performance of participating systems in the ELCardioCC 2026}
\centering
\begin{tabular}{|c|l|c|c|c|}
\hline
\textbf{Team} & \textbf{System} & \textbf{Recall} & \textbf{Precision} & \textbf{Micro-F1} \\
\hline
\multirow{1}{*}{stanimeros} 
  & ensemble\_metaheuristic\_p4\_winner... 
  & \textbf{0.850979} & 0.882995 & \textbf{0.866691} \\
\hline
\multirow{1}{*}{Georgios\_1} 
  & Baymax\_submission\_v1 
  & 0.865845 & 0.861783 & 0.863809 \\
\hline
\multirow{1}{*}{LSI\_UNED} 
  & test\_set\_NER\_greekuncased\_greek... 
  & 0.818709 & 0.841595 & 0.829994 \\
\hline
\multirow{1}{*}{alikibliona} 
  & submission\_A\_reranker 
  & 0.818347 & 0.824626 & 0.821474 \\
\hline
\multirow{1}{*}{elcardiocc} 
  & baseline\_bert\_base\_uncased\_greek 
  & 0.659173 & \textbf{0.903579} & 0.762264 \\
\hline
\multirow{1}{*}{Pakakisd} 
  & max\_micro\_curriculum 
  & 0.809282 & 0.720000 & 0.762035 \\
\hline
\multirow{1}{*}{fernandogd97} 
  & Greek\_ClinLinker-P\_greek\_bert\_DA.. 
  & 0.791516 & 0.734028 & 0.761689 \\
\hline
\end{tabular}
\label{tab:elcard}
\end{table}

The results show a relatively competitive top tier, with stanimeros achieving the best overall performance (F1 = 0.8667) using an ensemble-based approach that also yields the highest precision (0.8830), indicating strong balance between correct predictions and error control. Very closely behind, Georgios\_1 reaches an F1 of 0.8638, suggesting that different high-quality modeling strategies converge in performance at the top end. A mid-performing cluster includes LSI\_UNED and alikibliona, with F1 scores around 0.83–0.82, reflecting solid but less optimized systems. Below them, several systems (elcardiocc, Pakakisd, and fernandogd97) converge around 0.76 F1, but with notably different trade-offs: for example, the baseline BERT system prioritizes precision (0.9036) at the cost of recall, whereas others such as max\_micro\_curriculum favor recall more strongly. Overall, ensemble and correction-heavy methods clearly dominate the leaderboard, while simpler or single-model approaches tend to trade off precision and recall more unevenly, leading to lower or less balanced F1 scores.


\subsection{Task GutBrainIE}
Submitted runs were evaluated using micro- and macro-averaged precision, recall, and F1-score, with micro-F1 used as the reference measure for the leaderboards since it is better suited when classes are imbalanced. 

We adopted the same baseline architecture used in the previous edition, employing a fine-tuned NuNER model for NER and a fine-tuned ATLOP model for the RE subtasks \cite{bogdanov2024nuner,zhou2021atlop}. 
To support NERD and C-RE, we added a NEL module combining exact matching from the annotated training data with PubMedBERT-based semantic similarity over URI definitions for unmatched entities \cite{pubmedbert}.

Tables \ref{tab:gutbrainie_T611_scores}-\ref{tab:gutbrainie_T622_scores} show each team’s top run beating the baseline for NER, NERD, M-RE, and C-RE, respectively. 
NER obtained the highest scores, with a top micro-$F_1$ of $0.87$, confirming the effectiveness of biomedical pretrained transformers for exact-span entity recognition. 
Performances drop substantially in NERD, where the best micro-$F_1$ is $0.69$, showing that concept linking introduces significant challenges. 
A similar drop can be observed across RE subtasks: M-RE reaches $0.61$ micro-$F_1$, while C-RE drops to $0.35$. 
These results show that, although concept-level evaluation is not impacted by surface-form variations, it is highly sensitive to errors accumulated across entity recognition and linking, leading to a performance drop of over $40\%$ from M-RE to C-RE.

\begin{table}[!htb]
    \centering
    \caption{GutBrainIE performance metrics of each team's top run beating the baseline for NER. The best result is in bold, the second-best is underlined (micro-averaged).}
    \label{tab:gutbrainie_T611_scores}
    \begin{tabular}{L{0.25\linewidth}L{0.30\linewidth}M{0.15\linewidth}M{0.1\linewidth}M{0.1\linewidth}}
        \hline
        \textbf{Team ID} & \textbf{Run name} & \textbf{ Precision } & \textbf{ Recall } & \textbf{  F1  } \\
        \hline
        TWIX \cite{team_TWIX_gbie2026} & merged6 & \textbf{0.9548} & 0.8058 & \textbf{0.8740} \\
        NightSun \cite{team_NightSun_gbie2026} & run21 & \underline{0.8756} & 0.8110 & \underline{0.8420} \\
        Graphwise \cite{team_GraphWise_gbie2026} & 18 & 0.8613 & 0.8169 & 0.8385 \\
        TEXA \cite{team_TEXA_gbie2026} & 1 & 0.7980 & \underline{0.8302} & 0.8138 \\
        unibuc-bionlp-bp & 1 & 0.7816 & \textbf{0.8425} & 0.8109 \\
        SMTE \cite{team_SMTE_gbie2026} & R1 & 0.8109 & 0.7932 & 0.8019 \\
        GetGut@AAU \cite{team_GetGut_gbie2026} & 1 & 0.8011 & 0.8017 & 0.8014 \\
        NLP-DE & 1 & 0.7847 & 0.8173 & 0.8007 \\
        \hline
        BASELINE & NuNerZero-Finetuned & 0.7782 & 0.8221 & 0.7996 \\
        \hline
    \end{tabular}
\end{table}

\begin{table}[!htb]
    \centering
    \caption{GutBrainIE performance metrics of each team's top run beating the baseline for NERD. The best result is in bold, the second-best is underlined (micro-averaged).}
    \label{tab:gutbrainie_T612_scores}
    \begin{tabular}{L{0.25\linewidth}L{0.30\linewidth}M{0.15\linewidth}M{0.1\linewidth}M{0.1\linewidth}}
        \hline
        \textbf{Team ID} & \textbf{Run name} & \textbf{ Precision } & \textbf{ Recall } & \textbf{  F1  } \\
        \hline
        TWIX \cite{team_TWIX_gbie2026} & 24merged6 & \underline{0.7527} & \textbf{0.6353} & \textbf{0.6890} \\
        NightSun \cite{team_NightSun_gbie2026} & run16 & 0.6290 & \underline{0.6053} & \underline{0.6169} \\
        Graphwise \cite{team_GraphWise_gbie2026} & 9 & 0.6217 & 0.5897 & 0.6053 \\
        TUGW \cite{team_ToGS_gbie2026} & EXP4SUB1EL & \textbf{0.7723} & 0.4437 & 0.5636 \\
        MindGut link & BiomedBERT & 0.5738 & 0.5330 & 0.5527 \\
        GetGut@AAU \cite{team_GetGut_gbie2026} & 1 & 0.5515 & 0.5519 & 0.5517 \\
        GBA & 5 & 0.4385 & 0.5263 & 0.4784 \\
        TEXA \cite{team_TEXA_gbie2026} & 1 & 0.4581 & 0.4766 & 0.4672 \\
        \hline
        BASELINE & GLiNER-3stage & 0.4281 & 0.4522 & 0.4398 \\
        \hline
    \end{tabular}
\end{table}

\begin{table}[!htb]
    \centering
    \caption{GutBrainIE performance metrics of each team's top run beating the baseline for M-RE. The best result is in bold, the second-best is underlined (micro-averaged).}
    \label{tab:gutbrainie_T621_scores}
    \begin{tabular}{L{0.25\linewidth}L{0.30\linewidth}M{0.15\linewidth}M{0.1\linewidth}M{0.1\linewidth}}
        \hline
        \textbf{Team ID} & \textbf{Run name} & \textbf{ Precision } & \textbf{ Recall } & \textbf{  F1  } \\
        \hline
        TWIX \cite{team_TWIX_gbie2026} & mre10 & \textbf{0.8996} & \textbf{0.4651} & \textbf{0.6132} \\
        NightSun \cite{team_NightSun_gbie2026} & run18 & 0.4459 & \underline{0.4593} & \underline{0.4525} \\
        SMTE \cite{team_SMTE_gbie2026} & R2 & 0.4476 & 0.3997 & 0.4223 \\
        GetGut@AAU \cite{team_GetGut_gbie2026} & 1 & \underline{0.4546} & 0.3658 & 0.4054 \\
        GutHub \cite{team_GutHub_gbie2026} & 1 & 0.4347 & 0.3754 & 0.4029 \\
        Graphwise \cite{team_GraphWise_gbie2026} & BGPT5515GEXP8 & 0.3517 & 0.4497 & 0.3947 \\
        \hline
        BASELINE & Atlop-finetuned & 0.4444 & 0.3453 & 0.3886 \\
        \hline
    \end{tabular}
\end{table}

\begin{table}[!htb]
    \centering
    \caption{Performance metrics of each team's top run beating the baseline for C-RE. The best result is in bold, the second-best is underlined (micro-averaged).}
    \label{tab:gutbrainie_T622_scores}
    \begin{tabular}{L{0.25\linewidth}L{0.30\linewidth}M{0.15\linewidth}M{0.1\linewidth}M{0.1\linewidth}}
        \hline
        \textbf{Team ID} & \textbf{Run name} & \textbf{ Precision } & \textbf{ Recall } & \textbf{  F1  } \\
        \hline
        TWIX \cite{team_TWIX_gbie2026} & 11mre12 & \textbf{0.4903} & 0.2691 & \textbf{0.3475} \\
        NightSun \cite{team_NightSun_gbie2026} & run18 & \underline{0.2484} & \underline{0.2798} & \underline{0.2632} \\
        Graphwise \cite{team_GraphWise_gbie2026} & BGPT5515GEXP4 & 0.2092 & \textbf{0.2963} & 0.2452 \\
        GetGut@AAU \cite{team_GetGut_gbie2026} & 1 & 0.2123 & 0.1926 & 0.2020 \\
        TEXA \cite{team_TEXA_gbie2026} & 1 & 0.1556 & 0.1383 & 0.1464 \\
        \hline
        BASELINE & Atlop-3stage & 0.1403 & 0.1292 & 0.1345 \\
        \hline
    \end{tabular}
\end{table}

\section{Conclusions}
\label{sec:conclusion}

This paper provides an overview of the fourteenth BioASQ challenge.
This year, BioASQ consisted of six tasks: 
(i) \textit{Task 14b} on biomedical semantic question answering. 
(ii) \textit{Task Synergy14} on question answering for developing biomedical topics.
(iii) \textit{Task MultiClinSum-2} on multilingual clinical summarization. 
(iv) \textit{Task BioNNE-R} on relation extraction between nested named entities in Russian and
English. 
(v) \textit{Task ELCardioCC} on clinical coding in cardiology.
(vi) \textit{Task GutBrainIE} on gut-brain interplay information extraction.

The results for task 14b indicate that the strongest participant systems achieved high scores, particularly for yes/no answer generation, even in Phase A+, where expert-selected relevant material was not provided, which is in alignment with task 13b. For list and factoid questions, performance remains less consistent, especially in Phase A+, leaving clear room for further improvement. For these question types, access to expert-selected relevant material appears to help systems produce higher-quality answers.
These results also underline the importance of Phase A, which evaluates automatic retrieval of relevant material and shows variability across batches and a performance drop in task 14b, where more questions by new experts were included, compared to task 13b. 
Participants applied a diverse range of retrieval and generation strategies, including traditional methods, LLM and agent-based frameworks, multi-stage and ensembling techniques, and domain-specific knowledge integration.
The Synergy14 results, consistent with earlier editions, suggest that current systems can support biomedical scientists seeking specialized information on developing problems, while still leaving room for further advances.

The MultiClinSum-2 task focuses on the automatic summarization of clinical case reports across 15 languages, providing a comprehensive multilingual benchmark and evaluation framework to foster the development of automatic summarization systems capable of processing diverse clinical narratives within different linguistic contexts. The task received a total of 9 participating teams, with the highest participation concentrated in languages with greater resource availability such as English, Spanish or Portuguese. Participant systems showed rich and diverse methodological approaches, ranging from prompt engineering and agentic inference pipelines augmented with external biomedical knowledge such as UMLS, to parameter-efficient fine-tuning strategies such as QLoRA or reinforcement learning via GRPO. A key addition of this second edition is the introduction of an enhanced evaluation framework combining lexical and semantic summarization metrics with LLM-as-judge approaches, under which participating systems achieved notable results across all sub-tracks on both assessment dimensions.

The BioNNE-R task evaluated the extraction of relations between nested named entities in Russian and English biomedical texts. Seven teams participated, most framing the task as pair classification with typed entity markers and fine-tuned transformers, alongside one document-level graph model and LLM-based systems. All submissions outperformed the multilingual baseline, but the best macro-F1 scores stayed near 0.50–0.53, showing that relation extraction over nested entities remains an open challenge.

The ELCardioCC task centered on extracting and classifying medical entities from Greek discharge letters, drawing participation from six teams who submitted a diverse set of systems. Overall, the results show a clear shift toward hybrid pipelines and ensemble-based systems that combine NER, entity linking, and ICD-10 classification to better capture clinical semantics. Strong performance is consistently achieved by models that integrate multiple strategies and leverage Greek-specific transformer backbones, especially when combined with optimization techniques to address class imbalance. While simpler BERT-based baselines remain competitive, they are generally outperformed by more complex ensemble and knowledge-enhanced approaches. This indicates that combining structured clinical knowledge with multi-model fusion is key to achieving robust and accurate ICD-10 coding in this task.

The GutBrainIE task, centered on information extraction for the gut-brain axis, challenged participants with Named Entity Recognition and increasingly fine-grained Relation Extraction subtasks.
Teams achieving the strongest performance employed supervised deep-learning strategies, combining pretrained biomedical language models with ensemble strategies. 
Only a few participants experimented with prompt-based or generative approaches; however, these generally obtained lower scores, confirming the need to develop specialized models to effectively extract complex entities and relations in a specific biomedical domain.  

Overall, BioASQ 14 brought together 87 distinct teams across established and newer tasks, with substantial participation in all six tracks. 
Several participating systems achieved competitive performance, and some improved over baselines or previous state-of-the-art results.
In line with earlier editions, BioASQ continues to push research forward in semantic indexing, question answering,  information extraction, and language processing in biomedicine, offering both long-running and newer tasks. 
The challenge first expanded beyond English biomedical literature with MESINESP~\cite{luis2020overview}, and it has continued to broaden its scope since then. In its fourteenth edition, BioASQ included MultiClinSum-2~\cite{multiclinsum2-overview-2026}, BioNNE-R~\cite{bionner-overview-2026}, ELCardioCC~\cite{BioASQ2026ElCardioCC}, and GutBrainIE~\cite{BioASQ2026taskGutBrainIE}, alongside the established 14b and Synergy14 tasks. 
As a result, BioASQ 14 covered fifteen languages, multiple document types (biomedical articles, clinical case reports, and discharge letters), and specialized biomedical domains such as gut-brain interaction and cardiology.

Future directions for BioASQ include further expanding QA benchmark datasets through community-driven processes, broadening the network of biomedical experts involved, and enlarging the range of resources used across BioASQ tasks. This includes additional document types, more languages, and further specialized subdomains within biomedicine.

\section*{Acknowledgments}
The fourteenth edition of BioASQ is sponsored by Ovid.
The MEDLINE/PubMed data resources considered in this work were accessed courtesy of the U.S. National Library of Medicine.
BioASQ is grateful to the CMU team for providing the \textit{exact answer} baselines for task 14b.
The work on the BioNNE-R task is an output of a research project (HSE-BR-2025-025) implemented as part of the Basic Research Program at HSE University.
This research was funded by the Ministerio de Ciencia e Innovación (MICINN) under project BARITONE (TED2021-129974B-C22). This work is also supported by the European Union’s Horizon Europe Co-ordination \& Support Action under Grant Agreement No 101080430 (AI4HF), as well as Grant Agreement No 101057849 (DataTool4Heartproject).
ELCardioCC has been partially supported by project MIS 5154714 of the National Recovery and Resilience Plan Greece 2.0 funded by the European Union under the NextGenerationEU Program.
The work on the GutBrainIE task was supported by the HEREDITARY Project, as part of the European Union's Horizon Europe research and innovation programme (GA 101137074). 
%
%
%
\section*{Disclosure of Interests}
The authors have no competing interests to declare.

\bibliographystyle{splncs04}
\bibliography{BioASQ14.bib}

\end{document}